\documentclass[a4paper, 10pt, conference]{ieeeconf}      

\IEEEoverridecommandlockouts                              

\usepackage{graphics} 
\usepackage{graphicx,import}
\usepackage{graphbox}
\usepackage{svg}
\usepackage{subfig}

\usepackage{amsmath,amssymb,amsfonts} 

\usepackage{tikz}                      
\usepackage{tikzscale}
\usepackage{relsize}

\usepackage{pgfplots}                  
\pgfplotsset{compat=newest}
\usepgfplotslibrary{external}

\usepackage{cite}
\usepackage{textcomp}
\usepackage{marvosym} 
\usepackage{multirow}

\newcommand\copyrighttext{%
	\footnotesize \copyright~2019 IEEE. Personal use of this material is permitted. Permission from IEEE must be obtained for all other uses, in any current or future media, including reprinting/republishing this material for advertising or promotional purposes, creating new collective works, for resale or redistribution to servers or lists, or reuse of any copyrighted component of this work in other works. DOI: 10.1109/ITSC.2019.8917000}%
\newcommand\copyrightnotice{%
\begin{tikzpicture}[remember picture,overlay]
\node[anchor=south,yshift=10pt] at (current page.south) {\fbox{\parbox{\dimexpr\textwidth-\fboxsep-\fboxrule\relax}{\copyrighttext}}};
\end{tikzpicture}%
}

\title{\LARGE \bf
2D Car Detection in Radar Data with PointNets
}

\author{Andreas Danzer, Thomas Griebel, Martin Bach, and Klaus Dietmayer
\thanks{A. Danzer, T. Griebel, M. Bach, and K. Dietmayer are with the Institute of Measurement, Control and Microtechnology, Ulm University, 89081 Ulm, Germany
        {\tt\small firstname.lastname@uni-ulm.de}}%
}

\graphicspath{{img/}}

\newlength\figH
\newlength\figW
\def\hlinewd#1{%
    \noalign{\ifnum0=`}\fi\hrule \@height #1 %
    \futurelet\reserved@a\@xhline} 

\begin{document}

\maketitle
\copyrightnotice
\thispagestyle{empty}
\pagestyle{empty}

\begin{abstract}

	For many automated driving functions, a highly accurate perception of the vehicle environment is a crucial prerequisite. Modern high-resolution radar sensors generate multiple radar targets per object, which makes these sensors particularly suitable for the 2D object detection task. This work presents an approach to detect 2D objects solely depending on sparse radar data using PointNets. In literature, only methods are presented so far which perform either object classification or bounding box estimation for objects. In contrast, this method facilitates a classification together with a bounding box estimation of objects using a single radar sensor. To this end, PointNets are adjusted for radar data performing 2D object classification with segmentation, and 2D bounding box regression in order to estimate an amodal 2D bounding box. The algorithm is evaluated using an automatically created dataset which consist of various realistic driving maneuvers. The results show the great potential of object detection in high-resolution radar data using PointNets.
\end{abstract}

\section{Introduction}
\label{sec:introduction}
For autonomous driving, the perception of vehicle surrounding is an important task. In particular, tracking of multiple objects using noisy data from sensors, such as radar, lidar and camera, is a crucial task. A standard approach is to preprocess received sensor data in order to generate object detections, which are used as input for a tracking algorithm. For camera and lidar sensors, various object detectors have already been developed to obtain object hypotheses in form of classified rectangular 2D or 3D bounding boxes. To the best of the authors' knowledge, no object detection method for radar has been presented in literature so far which performs a classification as well as a bounding box estimation. Although modern high-resolution radar sensors generate multiple detections per object, the received radar data is extremely sparse compared to lidar point clouds or camera images. For this reason, it is a challenging task to recognize different objects solely using radar data.

\begin{figure}[ht]
	\vspace*{0.3cm}
	\centering
	\includegraphics[width=\columnwidth]{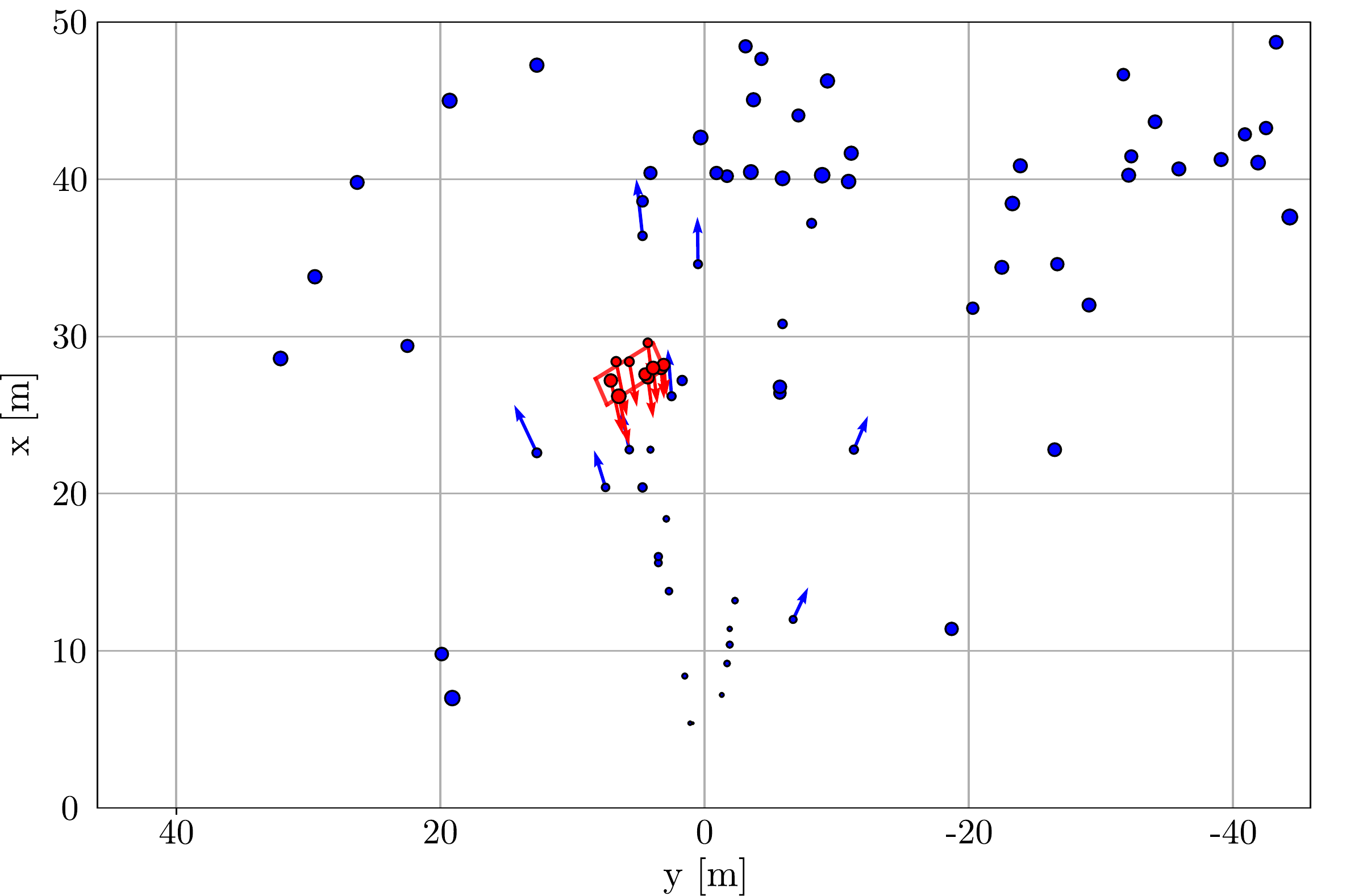}
	\caption{\textbf{2D object detection in radar data.} \, Radar point cloud with reflections belonging to a car (red) or clutter (blue). The length of arrows displays the Doppler velocity, the size of points represents the radar cross section value. The red box is a predicted amodal 2D bounding box.}
	\label{fig:radar_data}
\end{figure}

This contribution presents a method to detect objects in high-resolution radar data using a machine learning approach. As shown in Figure \ref{fig:radar_data}, radar data is represented as a point cloud, called radar target list, consisting of two spatial coordinates, ego motion compensated Doppler velocity and radar cross section (RCS) values. 
Since radar targets are represented as point cloud, it is desirable to use raw point clouds as input for a neural network. Existing approaches which process radar data use certain representation transformations in order to use neural networks, e.g. radar data is transformed into a grid map representation. However, PointNets \cite{qi2017pointnet, qi2017pointnetplusplus} make it possible to directly process point clouds and are suitable for this data format.
Frustum PointNets \cite{qi2018frustum} extend the concept of PointNets for the detection of objects by combining a 2D object detector with a 3D instance segmentation and 3D bounding box estimation. The proposed object detection method for radar data is based on the approach of Frustum PointNets to perform 2D object detection, i.e., object classification and segmentation of radar target lists together with a 2D bounding box regression. The 3D lidar point cloud used in \cite{qi2017pointnet, qi2017pointnetplusplus, qi2018frustum} is dense and even captured fine-grained structures of objects. Although radar data is sparse compared to lidar data, radar data contains strong features in form of Doppler and RCS information. For example, wheels of a vehicle generate notable Doppler velocities, and license plates result in targets with high RCS values. Another advantage of radar data is that it often contains reflections of an object part which is not directly visible, e.g., wheel houses at the opposite side of a vehicle. All these features can be very beneficial to classify and to segment radar targets, and above all to perform bounding box estimations of objects.

This work is structured as follows. Section \ref{sec:relatedwork} displays related work in the field of object classification and bounding box estimation using radar data. Furthermore, deep learning approaches on point clouds using PointNets are introduced. Section \ref{sec:problemstatement} provides a description of the problem that is solved by this contribution. Section \ref{sec:radarpointnets} presents the proposed method to detect 2D objects in radar data. Additionally, the automatically generated radar dataset is described, and the training process is explained. Section \ref{sec:experiments} shows results which are evaluated on real-world radar data.

\section{Related Work}
\label{sec:relatedwork}
\paragraph*{Object Classification in Radar Data}
For object classification in radar data, Heuel and Rohling \cite{heuel2011,heuel2012,heuel2013} present approaches with self-defined extracted features to recognize and classify pedestrians and vehicles. W\"ohler et al. \cite{woehler2017} extract stochastic features and use them as input for a random forest classifier and a long short-term memory network. In order to classify static objects, Lombacher et al. \cite{lombacher2016} accumulate raw radar data over time and transform them into a grid map representation \cite{werber2015}. Then, windows around potential objects are cut out and used as input for a deep neural network. Furthermore, Lombacher et al. \cite{lombacher2017_semanticradargrids} infer a semantic representation for stationary objects using radar data. To this end, a convolutional neural network is fed with an occupancy radar grid map. 

\paragraph*{Bounding Box Estimation in Radar Data}
Instead of classifying objects, another task is to estimate a bounding box, i.e., position, orientation and dimension of an object. Roos et al. \cite{roos2016} present an approach to estimate the orientation as well as the dimension of a vehicle using high-resolution radar. For this purpose, single measurements of two radars are collected and enhanced versions of orientated bounding box algorithms and the L-fit algorithm are applied. Schlichenmaier et al. \cite{schlichenmaier2016} show another approach to estimate bounding boxes using high-resolution radar data. For this reason, position and dimension of vehicles are estimated using a variant of the $k$-nearest-neighbors method. Furthermore, Schlichenmaier et al. \cite{schlichenmaier2017} present an algorithm to estimate bounding boxes representing vehicles using template matching. Especially in challenging scenarios, the templating matching algorithm outperforms the orientated bounding box methods. A disadvantage of the proposed method is that clutter points not belonging to the vehicle are however taken into account for the bounding box estimation.

\paragraph*{PointNets}
The input of most neural networks has to follow a regular structure, e.g., image grids or grid map representation. This requires that data such as point clouds or radar targets have to be transformed in a regular format before feeding them into a neural network. The PointNet architecture overcomes this constraint and supports point clouds as input. Qi et al. \cite{qi2017pointnet} present a 3D classification and a semantic segmentation of 3D lidar point clouds using PointNet. Since the PointNet architecture does not capture local structures induced by the metric space, the ability to capture details is limited. For this reason, Qi et al. \cite{qi2017pointnetplusplus} also propose a hierarchical neural network, called PointNet++, which applies PointNet recursively on small regions of the input point set. 
Schumann et al. \cite{schumann2018} use the same PointNet++ architecture for semantic segmentation on radar point clouds. For this purpose, the architecture is modified to handle point clouds with two spatial and two further feature dimensions. The radar data is accumulated over a time period of $500$\,ms to get a denser point cloud with more reflections per object. Subsequently, each radar target is classified among six different classes. Thus, only a semantic segmentation is performed but no semantic instance segmentation or bounding box estimation.
Utilizing image and lidar data, Qi et al. \cite{qi2018frustum} present the Frustum PointNets, to detect 3D objects. First, a 3D frustum point cloud including an object is extracted using a 2D bounding box from an image based object detector. Second, a 3D instance segmentation in the frustum is performed using a segmentation PointNet. Third, an amodal 3D bounding box is estimated using a regression PointNet. Hence, Frustum PointNets are the first method performing object detection including bounding box estimation using PointNets on unstructured data.

\section{Problem Statement}
\label{sec:problemstatement}
Given radar point clouds, the goal of the presented method is to detect objects, i.e., to classify and localize objects in 2D space (Figure \ref{fig:radar_data}). The radar point cloud is represented as a set of four-dimensional points ${\mathcal{P} = \left\{p_i \vert i = 1,\dots,n \right\}}$, where $n \in \mathbb{N}$ denotes the number of radar targets. Furthermore, each point $p_i = (x,y,\tilde{v}_r,\sigma)$ contains $(x,y)$-coordinates, ego motion compensated Doppler velocity $\tilde{v}_r$, and radar cross section values $\sigma$. The radar data are generated from one measurement cycle of a single radar and are not accumulated over time.

For object classification, this contribution distinguishes between two classes, \textit{car} and \textit{clutter}. 
Additionally, for radar targets that are assigned to the class car, an amodal 2D bounding is predicted, i.e., even if only parts are captured by the radar sensor, the entire object is estimated. The 2D bounding box is described by its center $(x_c, y_c)$, its heading angle $\theta$ in the $xy$-plane and its size containing length $l$ and width $w$.

\section{2D Object Detection with PointNets}
\label{sec:radarpointnets}
\begin{figure*}[ht]
	\centering
	\includegraphics[width=0.95\textwidth]{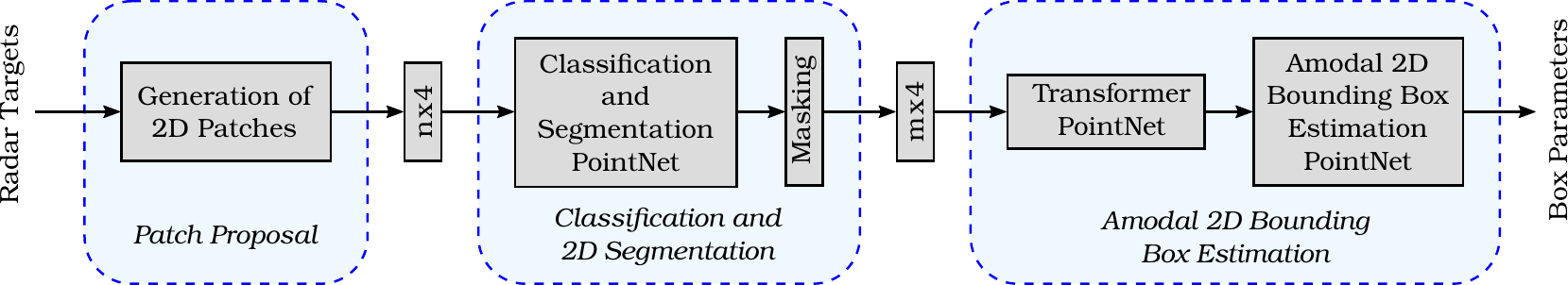}
	\caption{\textbf{2D object detection in radar data with PointNets.} \, First, a patch proposal determines multiple region of interests, called patches, using the entire radar target list. Second, a classification and segmentation network classifies these patches. Subsequently, each of the $n$ radar targets are classified to get an instance segmentation. Finally, a regression network estimates an amodal 2D bounding box for objects using the $m$ segmented car radar targets.}
	\label{fig:radarpointnets}
\end{figure*}

An overview of the proposed 2D object detection system using radar data is shown in Figure \ref{fig:radarpointnets}. This sections introduces the three major modules of the system: patch proposal, classification and segmentation, and amodal 2D bounding box estimation.

\subsection{Patch Proposal}
\label{sec:patch_proposal}

The patch proposal divides the radar point cloud into regions of interest. For this purpose, a patch with specific length and width is determined around each radar target. The length and width of the patch must be selected in such a way that it comprises the entire object of interest, here a car. Additionally, it is important that each patch contains enough radar targets to distinguish between car and clutter patches in the classification as well as car and clutter radar targets in the segmentation step. 
The patch proposal generates multiple patches containing the same object. As a result, the final 2D object detector provides multiple hypotheses for a single object. 
This behavior is desirable because the object tracking system in the further processing chain for environmental perception deals with multiple hypotheses per object. Note that the tracking system is not part of this work. 
As described in \cite{qi2018frustum}, the patches are normalized to a center view which ensures rotation-invariance of the algorithm. Finally, all radar targets within a patch are forwarded to the classification and segmentation network.

\subsection{Classification and Segmentation}

The classification and object segmentation module consists of a network which classifies each patch and segments all radar targets inside the patch. For this purpose, the entire patches are considered using the classification network to distinguish between \textit{car} and \textit{clutter} patches. For car patches, the segmentation network predicts a probability score for each radar target which indicates the probability of radar targets belonging to a car. In the masking step, radar targets which are classified as car targets are extracted. As presented in \cite{qi2018frustum}, coordinates of the segmented radar targets are normalized to ensure translational invariance of the algorithm. 

Note that the classification and segmentation module can easily be extended to multiple classes. For this purpose, the patch is classified as a certain class and, consequently the predicted classification information is used for the segmentation step.

\subsection{Amodal 2D Bounding Box Estimation}

After the segmentation of object points, this module estimates an associated amodal 2D bounding box. First, a light-weight regression PointNet, called Transformer PointNet (T-Net), estimates the center of the amodal bounding box and transforms radar targets into a local coordinate system relative to the predicted center. This step and the T-Net architecture are described in detail in \cite{qi2018frustum}. The transformation using T-Net is reasonable, because regarding the viewing angle, the centroid of segmented points can differ to the true center of the amodal bounding box.

For the 2D bounding box estimation, a box regression PointNet, which is conceptually the same as proposed in \cite{qi2018frustum}, is used. The regression network predicts parameters of a 2D bounding box, i.e., its center $(x_c, y_c)$, its heading angle $\theta$ and its size $(l,w)$. For the box center estimation, residual based 2D localization of \cite{qi2018frustum} is performed. Heading angle and size of bounding box is predicted using a combination of a classification and a segmentation approach as explained in \cite{qi2018frustum}. More precisely, for size estimation, predefined size templates are incorporated for the classification task. Furthermore, residual values regarding those categories are predicted.

In case of multiple classes, the box estimation network also uses the classification information for the bounding box regression. Therefore, the size templates have to be extended by additional classes, e.g., \textit{pedestrians} or \textit{cyclists}.

\subsection{Network Architectures}

\begin{figure*}[ht]
\centering
\subfloat{
	{\includegraphics[align=t,width=0.82\textwidth]{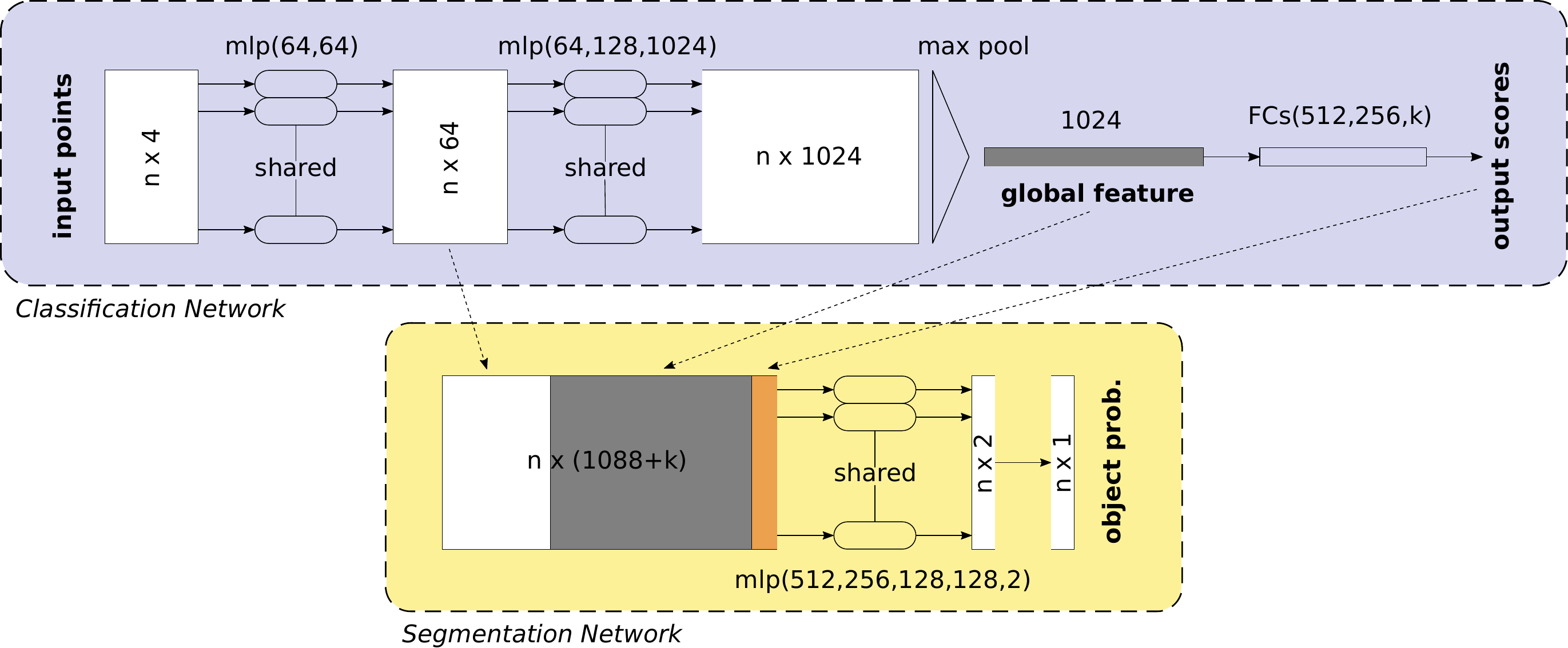}}
}\hfill
\subfloat{
	{\includegraphics[align=t,width=0.65\textwidth]{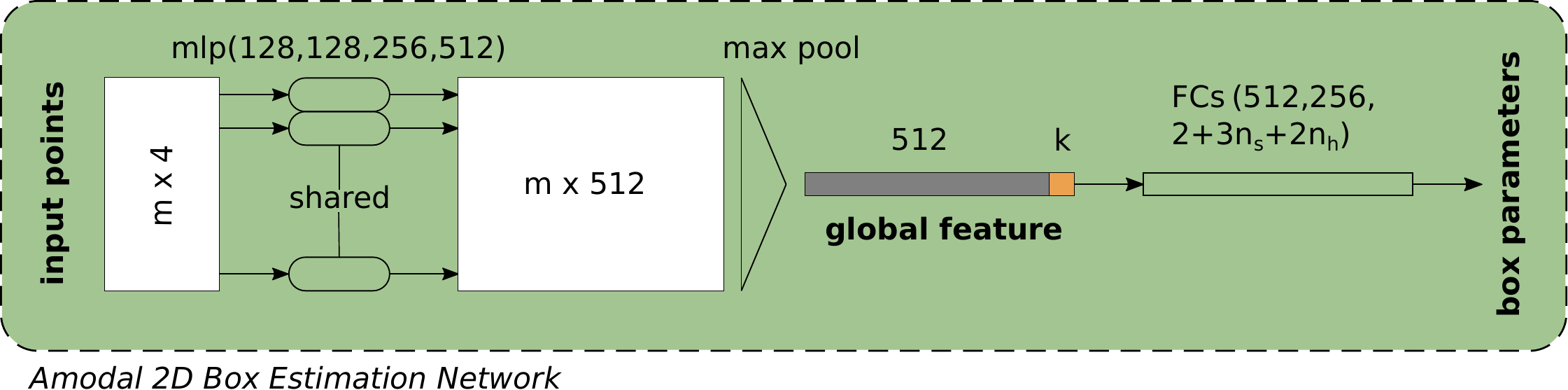}}
}
\caption{\textbf{Network architectures for 2D object detection in radar data with PointNets.} \, The model is based on PointNet. Input points are a list of radar targets. To get global features in the classification network, the model consists of a multi-layer perceptron (mlp) and max pooling layers. After fully connected (FC) layers, classification scores for the whole point cloud are obtained. In the segmentation network, local and global features are combined with output scores from the classification, the model applies mlp to obtain object probabilities. In the box estimation network, it contains mlp and max pooling layers to get global features. Features are combined with classification scores and are forwarded to FC layers to determine box parameters.}
\label{fig:network_architectures}
\end{figure*}

For the object detecion task in radar data, the network architecture is based on the concepts of PointNet \cite{qi2017pointnet} and Frustum PointNets \cite{qi2018frustum}. Figure \ref{fig:network_architectures} shows the network architecture consisting of classification, segmentation and 2D bounding box regression network. For the classification and segmentation network the architecture is conceptually similar to \cite{qi2017pointnet}. The network for amodal 2D bounding box estimation is the same as proposed in \cite{qi2018frustum}. 
In this work, the input of the classification and bounding box regression network is radar data. For this reason, the input regarding original PointNet is extended for radar target lists. For classification and segmentation network as well as bounding box regression network, the radar targets are represented as set of four-dimensional points containing 2D spatial data, ego motion compensated Doppler velocity and RCS information. For the classification and segmentation network, the input is a radar target list with $n$ points of a patch. Then, the segmented radar target list with $m$ points belonging to an object is fed into the 2D bounding box estimation network.

\subsection{Dataset}

For training and testing of the proposed 2D object detection method in radar data, a dataset with real-world radar data was created on a test track. Two test vehicles of Ulm University \cite{kunz2015} were used, an ego vehicle and a target vehicle. 
The ego vehicle is equipped with two ARS 408-21 Premium radar sensors from Continental that are installed on the front corners of the vehicle. Note in this work, only the radar sensor mounted on the front left corner is used. Both vehicles are equipped with an Automotive Dynamic Motion Analyzer (ADMA) sensor unit from GeneSys and a highly accurate Differential Global Positioning System (DGPS). The target vehicle is a Mercedes E-Class station wagon S212. The ground truth data of the target vehicle is generated using ADMA data and represented as a bounding box comprising position, orientation, and dimension.

At the time of recording the dataset, there were different weather conditions. The dataset consists of eleven different driving maneuvers, e.g., driving circles and figure eights, driving in front or towards the ego vehicle, as well as passing maneuvers. The idea is to cover as many maneuvers under training conditions, which can be transferred to real-world traffic scenarios, as possible.

The labeling process generates automatically annotated radar targets by using the ground truth bounding box of the target vehicle as reference. Each radar target inside the reference bounding box is marked as car target, and all the others as clutter targets. Since radar measurements are typically noisy, targets close to the target vehicle can belong to the car class. Therefore, the ground truth bounding box is extended by $0.35$\,m in length and width. The big advantage of the automatic label procedure is that a large amount of data can be annotated with minimal effort.

After patch proposal, i.e., generating patches around each radar target, patches are annotated as car if the radar target which defines the patch (the radar target in the center of the patch) belongs to the car class. Otherwise the patch is regarded as clutter patch. Since a certain number of radar targets are necessary to assign a class as well as to estimate a bounding box, the dataset is created under certain conditions. Hence, each car patch has to comprise at least $2$ radar targets belonging to the car class, and each clutter patch  has to comprise at least $16$ radar targets belonging to the clutter class. Since there are much more examples for clutter patches, the resulting distribution of the annotated patches is imbalanced, more precisely only $4.77 \%$ of the patches belongs to the car class.

For training, validation, and testing purpose, the dataset is divided into three parts: training, validation and testing data. The training data is used to train the model. The evaluation of the model during training process uses the validation data. The validation data consists of same driving maneuvers as the training data. For example, if a maneuver was driven five times, then three of them are added to the training data and the rest to the validation data. This ensures that all trained driving maneuvers are covered within the validation data. The testing data is used to show the generalization ability of models after the training process. The testing data consists of different driving maneuvers which differ from the trained maneuvers. As a result, the testing data is completely disjoint from the training and the validation data. In total, $61.68 \%$ of the patches are used for training, $19.62 \%$ for validation and $18.70 \%$ for testing.

\subsection{Training}

As proposed in \cite{qi2018frustum}, training is performed with a multi-task loss to optimize classification and segmentation PointNet, T-Net and amodal 2D bounding box estimation PointNet simultaneously. Since this work uses a patch classification before segmentation, the multi-task loss is extended by the classification part as proposed in \cite{qi2017pointnet}. As a result, the multi-task loss is defined as

\begin{align}
	L_{multi-task} = & w_{cls} \, L_{cls} + w_{seg} \, L_{seg} +  w_{box} \, ( L_{c1-reg} \nonumber \\
					 & + L_{c2-reg} + L_{h-cls} +  L_{h-reg} + L_{s-cls} \nonumber \\
					 & + L_{s-reg} + w_{corner} \, L_{corner} ).
	\label{eq:multi_task_loss}
\end{align}

In Equation (\ref{eq:multi_task_loss}), $L_{cls}$ denotes the loss for patch classification and $ L_{seg}$ for segmentation of radar targets in the patch. Both losses can be weighted by the parameters $w_{cls}$ and $w_{seg}$ respectively. Since the distribution of car and clutter patches in the dataset is imbalanced, during training process the weights for $w_{cls}$ and $w_{seg}$ are chosen to be higher for car patches. All other losses are used for bounding box regression. Here, $L_{c1-reg}$ is for residual based center regression of T-net and $L_{c2-reg}$ for center regression of amodal box estimation network. Furthermore, $L_{h-cls}$ and $L_{h-reg}$ are losses for heading angle estimation, while $L_{s-cls}$ and $L_{s-reg}$ are for size estimation of the bounding box. The corner loss $L_{corner}$, weighted by $w_{corner}$, is a novel regularization loss proposed in \cite{qi2018frustum} for jointly optimizing center, size and heading angle of the bounding box. This loss is constructed to ensure a good 2D bounding box estimation under the intersection over union metric. The parameter $w_{box}$ weights the bounding box estimation. If a patch is classified as clutter patch, $w_{box}$ is set to zero with the result that no bounding box estimation is performed. Moreover, softmax with cross-entropy loss is used for the classification and the segmentation task, and smooth-$l_1$ (huber) loss is used for loss calculations of the regression task.

To guarantee a fixed number of input points, sampling is performed. For the classification and segmentation, up to $48$ radar targets from each patch are drawn. Since the number of radar targets generated by car class is sparse, the sampling process during training considers all car targets and only samples clutter points. When testing the object detector, radar targets are sampled independent regarding the class. For the amodal 2D bounding box estimation, up to $32$ points are randomly sampled from the segmented radar targets. 

Data augmentation is a helpful concept to avoid model overfitting. 
Hence, for all radar targets of a patch, data augmentation is applied during training. First, spatial information in a radar patch are perturbed by randomly shifting all radar targets uniformly regarding $x$ and $y$ direction. Second, for car targets ego motion compensated Doppler velocity is perturbed using random Gaussian noise with zero mean and standard deviation of $0.2$. Third, for perturbing RCS value of car targets, random noise of a Gaussian distribution with zero mean and standard deviation of one is used.

Training of the model uses Adam optimizer and performs batch normalization for all layers excluding the last classification and regression layers. The initial learning of both models is chosen to be $0.0001$. Further, batch  size for for training is chosen to be $32$. Training is performed for $11$ epochs on a single \textit{NVIDIA GeForce GTX 1070} GPU.

The model is trained on the full dataset containing patches with at least $2$ car targets and $16$ clutter targets per patch. Furthermore, the weights for multi-task loss during training are heuristically chosen to be $w_{cls}=2$, $w_{seg}=2$ for car patches and $w_{cls}=1$, $w_{seg}=1$ for clutter patches, and $w_{box}=1$, $w_{corner}=10$ for the bounding box estimation. The weights for the bounding box estimation loss are chosen to be significant higher, since the optimization of the bounding box regression using sparse radar targets point cloud is more complex than the classification and segmentation task. This is due to the fact that Doppler velocities and RCS values are strong features, because RCS and Doppler velocities of car and clutter radar targets differ significantly.

\section{Experiments}
\label{sec:experiments}
This section presents the results of the proposed 2D object detector in radar data. For this purpose, classification and segmentation as well as 2D bounding box estimation is evaluated. Furthermore, results and restrictions are discussed.

\subsection{Evaluation}

Since the proposed 2D object detection method contains classification, segmentation, and bounding box estimation, the performance of these three modules will be evaluated.
For classification and segmentation, accuracy and $F_1$ score is used. For bounding box regression, the performance is measured by Intersection over Union (IoU) metric.

The $F_1$ score is the harmonic mean between precision $P$ and recall $R$ and is given by
\begin{equation}
	F_1 = \dfrac{2 \cdot P \cdot R}{P + R}.
\end{equation}

The IoU compares a predicted bounding box $b_{pred}$ with the ground truth bounding box $b_{gt}$ and is defined as
\begin{equation*}
\text{IoU} = \frac{| b_{gt} \cap b_{pred} |}{| b_{gt} \cup b_{pred} |},
\end{equation*}
where $| \cdot |$ measures the area of the underlying set. If the ground truth and the predicted bounding box are almost the same, IoU score tends to be close to one. If the two bounding boxes do not overlap at all, IoU score will be zero.

\begin{table*}[t!]
	\vspace*{0.3cm}
	\centering
	\caption{\textbf{Results for 2D object detection in radar data.} \, The object detector is evaluated on the test set. Accuracy and $F_1$ score are evaluated for classification and segmentation. IoU evaluates the 2D bounding box estimation by using mean IoU (mIoU) and ratio of IoUs with a threshold of $0.7$. The entire test data set as well as single driving maneuvers are considered.} \label{tab:object_detection_results}
	\begin{tabular}{l||cc|cc|cc}
		\hlinewd{1pt}
		\multirow{2}{*}{Test data} & \multicolumn{2}{c|}{Classification} & \multicolumn{2}{c|}{Segmentation} & \multicolumn{2}{c}{2D Bounding Box} \\
		&  Accuracy & $F_1$ score  & Accuracy & $F_1$ score & mIoU & IoU $(\geq0.7)$\\\hline
		entire test dataset& $96.07 \%$ & $64.90 \%$ & $82.64 \%$ & $86.58 \%$ & $0.64$ & $59.43 \%$ \\\hline
		target vehicle circling around standing ego vehicle & $91.90 \%$ & $32.68 \%$ & $93.30 \%$ & $50.33 \%$ & $0.56$ & $50.70 \%$ \\ 
		target vehicle crossing standing ego vehicle & $89.51 \%$ & $21.90 \%$ & $91.56 \%$ & $40.81 \%$ & $0.54$ & $43.96 \%$ \\ 
		target vehicle passing standing ego vehicle & $93.73 \%$ & $41.29 \%$ & $95.09 \%$ & $52.76 \%$ & $0.56$ & $51.47 \%$ \\ 
		ego vehicle driving behind target vehicle & $95.02 \%$ & $41.89 \%$ & $87.84 \%$ & $69.61 \%$ & $0.71$ & $66.08 \%$ \\ 
		vehicles driving behind each other with overtaking & $97.52 \%$ & $50.38 \%$ & $80.46 \%$ & $77.89 \%$ & $0.73$ & $72.09 \%$ \\ 
		\hlinewd{1pt}
	\end{tabular}
\end{table*}

Testing was performed on the full dataset, where the number of car and clutter targets per patch are at least $2$ and $16$ respectively. The length and the width of the patches (Section \ref{sec:patch_proposal}) are chosen to be $10$\,m, which guarantees that a patch comprises an entire vehicle at any time. Further, single driving maneuvers are evaluated. Therefore, the radar object detection processes all measured radar targets from a sensor's measurement cycle. Hence, the condition that each car and clutter patch consists of at least $2$ car and $16$ clutter radar targets respectively, is not given. Table \ref{tab:object_detection_results} shows the results of the object detection in radar data. 
The proposed object detector shows promising results regarding classification and segmentation accuracy as well as bounding box regression.
The object detector shows best results regarding 2D bounding box estimation in maneuvers, where the ego vehicle is following the target vehicle with and without overtaking.
Considering inference time, the proposed object detector takes $2.9$\,ms for prediction (classification and segmentation together with amodal 2D bounding box estimation) per patch.

\subsection{Discussion}
\label{sec:discussion}
Generally, the results of the 2D object detector in radar data are promising. However, it is important to note that the presented dataset is limited. The dataset contains only one object per radar measurement cycle. A further restriction is that the object has always the same ground truth size, such that the result of size estimation must not be considered in more detail. Nevertheless, with the proposed method provides the mechanisms to deal with different sizes, but their effect has to be evaluated in future work. Same argumentation holds for the detection of multiple classes. The mechanisms to recognize multiple classes is provided, but the current dataset does not include objects with different classes. For future research, the dataset will be extended to multiple objects and multiple classes.

Another important point to note is that in further work, the object detector for radar data will be a preprocessing module for a multi-object tracking system which fuses data of multiple radar sensors. This is the main reason why no accumulation of multiple radar measurements is used in order to get point clouds that are more dense. The fusion of radar measurements with different time stamps, as well as measurements from multiple radar sensors, can be processed with a sophisticated object tracking system. Additionally, the object tracker probabilistically models multiple hypotheses per object. So it is desirable that the object detector generates multiple hypotheses which is guaranteed by the patch proposal module. Since one radar sensor is only able to provide sparse data per measurement cycle, the proposed object detector will naturally generate misdetections or clutter measurements. However, a multi-object tracker such as the Labeled Multi-Bernoulli Filter \cite{reuter2014} is able to handle this.

\section{Conclusion}
\label{sec:conclusion}
This contribution has proposed an approach to detect 2D objects hypotheses in sparse radar data. The object detector performs an object classification and a 2D segmentation, together with an amodal 2D bounding box estimation using variants of PointNets. 
Although, in this work only one object class was considered, the results are promising and the proposed approach should be further investigated. In future work, the dataset will be extended by several classes to investigate the performance of the object detector for multiple classes. Additionally, objects in the dataset belonging to the same class will have different sizes to examine the impact of object dimension.

\section*{Acknowledgment}
This work was performed as part of the second author's research within a thesis to obtain the Master of Science. 

This research is accomplished within the project “UNICARagil” (FKZ 16EMO0290). We acknowledge the financial support for the project by the Federal Ministry of Education and Research of Germany (BMBF).

\bibliographystyle{IEEEtran}

\bibliography{IEEEabrv,references}

\end{document}